\RequirePackage{fix-cm}
\documentclass[twocolumn]{svjour3}          
\smartqed  
\usepackage{graphicx}
\usepackage{amsmath,amssymb,amsfonts}
\usepackage{amsmath}
\usepackage{mathrsfs}
\usepackage{xcolor}
\usepackage[numbers]{natbib}
\usepackage{ulem}
\usepackage[shortcuts]{extdash}
\usepackage[utf8]{inputenc}
\begin{document}

\title{Pressure Field Reconstruction with SIREN}
\subtitle{A Mesh-Free Approach for Image Velocimetry in Complex Noisy Environments}


\author{Renato F. Miotto \and
        William R. Wolf \and
        Fernando Zigunov
}


\institute{R. F. Miotto \at
          Department of Energy, School of Mechanical Engineering - Universidade Estadual de Campinas - R. Mendeleyev 200, Campinas, Brazil \\
          \email{miotto@fem.unicamp.br}           
        \and
        W. R. Wolf \at
            Department of Energy, School of Mechanical Engineering - Universidade Estadual de Campinas - R. Mendeleyev 200, Campinas, Brazil 
        \and
        F. Zigunov \at
            Department of Mechanical and Aerospace Engineering - Syracuse University - 263 Link Hall, Syracuse, NY, USA
}

\date{Received: date / Accepted: date}

\maketitle


\abstract{
This work presents a novel approach for pressure field reconstruction from image velocimetry data using SIREN (Sinusoidal Representation Network), emphasizing its effectiveness as an implicit neural representation in noisy environments and its mesh-free nature. While we briefly assess two recently proposed methods — one-shot matrix-omnidirectional integration (OS\-/MODI) and Green's function integral (GFI) — the primary focus is on the advantages of the SIREN approach. The OS\-/MODI technique performs well in noise-free conditions and with structured meshes but struggles when applied to unstructured meshes with high aspect ratio. Similarly, the GFI method encounters difficulties due to singularities inherent from the Newtonian kernel. In contrast, the proposed SIREN approach is a mesh-free method that directly reconstructs the pressure field, bypassing the need for an intrinsic grid connectivity and, hence, avoiding the challenges associated with ill\-/conditioned cells and unstructured meshes. This provides a distinct advantage over traditional mesh-based methods. Moreover, it is shown that changes in the architecture of the SIREN can be used to filter out inherent noise from velocimetry data. This work positions SIREN as a robust and versatile solution for pressure reconstruction, particularly in noisy environments characterized by the absence of mesh structure, opening new avenues for innovative applications in this field.
}




\maketitle


\section{Introduction}
\label{sec:intro}

Recent advancements in image-based experimental techniques allowed the acquisition of velocity measurements with high temporal and spatial resolution \citep{Beresh2021}. This has opened the path to obtaining instantaneous fields of different quantities of interest in a non-intrusive manner by processing experimental data in combination with analysis of fluid flow equations. Pressure plays an example of such a quantity of interest given its usefulness for engineering design  \citep{MARTINS2022105391}, fluid flow analysis \citep{Percin2017, McClure2017}, noise generation \citep{Carter2023} and aerodynamic response of impulsive flows \citep{Savelli2023}. In this sense, different pressure reconstruction methods have been developed on top of particle image velocimetry (PIV) and Lagrangian particle tracking (LPT) data.

For incompressible flows, pressure is obtained through spatial integration of the momentum equation
%
%
%
\begin{equation}
    \nabla p = 
    - \rho \frac{D\textbf{u}}{D t} + \mu \nabla^2 \textbf{u}
    \ \mbox{,}
    \label{eq:momentum}
\end{equation}
%
where $D\textbf{u} / D t$ is the material acceleration, $\rho$ is the fluid density and $\mu$ is the dynamic viscosity. To perform this integration, several classes of methods have been proposed, such as least-square reconstruction \cite{Jeon2018}, direct line integration \citep{Dabiri2014, Liu2006, Liu2020}, solution of a pressure Poisson equation \citep{Charonko2010, Goushcha2023, Wang_2023}, non-parametric kernel-based probabilistic models \citep{you2023}, and physics-informed neural networks \citep{Calicchia2023, Fan2023}. In addition to the integration methods themselves, an extensive research has been carried out on the experimental determination of the material acceleration (see \citep{Jensen2004, vanGent2017, Violato2011} for instance), which is usually the term that dominates the pressure gradient balance at high Reynolds numbers. The diversity of pressure estimation methods in PIV and LPT arises from limitations inherent to optical-based flow velocimetry setups, such as spatial or temporal resolution, field of view, or accessibility. Various techniques have been proposed to overcome these constraints and obtain meaningful pressure data. While some methods prioritize accuracy, they may demand greater computational resources or expertise \citep{Liu2020}, whereas simpler approaches may compromise accuracy not only for ease of implementation but also to reduce computational costs \citep{Dabiri2014, VanderKindere2019}.

While our primary focus in this work is on estimating pressure, specifically through the integration of Eq. \ref{eq:momentum}, it is essential to acknowledge the significance of accurately evaluating the material derivative. The work of \citet{Pan2016} highlights the importance of this evaluation, demonstrating that the propagated error from a Poisson solver is contingent upon factors such as boundary conditions, domain shape (size and aspect ratio), and the error in the pressure gradient at the boundaries and flow field. These findings are aligned with the benchmark study conducted by \citet{Charonko2010}, that compared various PIV-based methods for pressure calculation through both numerical simulations and experiments and revealed the sensitivity of pressure calculation performance to numerous factors. In this context, the methods presented by \citet{Azijli2016} and \citet{Zhang2022} offer practical approaches for assessing the uncertainty associated with reconstructed pressure fields.

When the material acceleration and boundary conditions are erroneous, a popular strategy is to average multiple pressure calculations along different integral paths. This approach capitalizes on the scalar nature of the pressure field, ensuring that the integrated pressure value at any given location must be independent of the integral path. A well-known implementation of such approach is the omni-directional integration (ODI) method, initially proposed by \citet{Liu2006}. This class of method, in particular the rotating parallel ray variation \citep{Liu2016}, was regarded as highly effective in removing a substantial portion of random errors, thereby exhibiting lower sensitivity to measurement errors compared to Poisson-based solvers \citep{Charonko2010, Liu2020}. Insights from \citet{Liu2020} suggest that this advantage stems from its improved handling of errors at domain boundaries, which seemed to render the imposition of Dirichlet boundary conditions upon the final convergence of iterations.

\citet{Zigunov2024MODI, zigunov2024oneshot} expanded upon the rotating parallel ray ODI technique, achieving omni-directional integration at high resolution and with a single iteration. This was accomplished by recasting the problem as an iterative matrix inversion task in the limit of infinite iterations. Besides reaching a computational speedup of $\sim 10^6$ compared to the traditional technique, which enables treatment of volumetric measurements, they established a connection between ODI and Poisson-based solvers. For a uniform grid, their implementation of the ODI (labeled one-shot matrix ODI, OS\-/MODI) degenerates to the finite volume discretization of the pressure Poisson equation (PPE), albeit with a boundary treatment that differs from a na\"ive Neumann boundary condition implementation (ghost volume). This new comprehension was recently formalized by \citet{Pryce2024, pryce2024revisit}, alongside a clarification regarding the treatment of boundary conditions in the ODI: they rigorously proved that by running a conjugate gradient (CG) method on ODI is equivalent to solving a PPE with Neumann boundary conditions by regularization, specifically pursuing a minimal norm solution of the resulting under-determined system. 

With this established connection between the two classes of methods, insights gained from previous studies on error propagation in the PPE \citep{Pan2016, Faiella2021, Nie2022} can be effectively utilized to enhance understanding of the reconstructed pressure field derived from the ODI - at least for the case of a uniform grid spacing. It also suggests that the divergent outcomes reported in the literature using ODI  (e.g., ODI exhibited significantly higher accuracy than PPE in studies by \citet{Charonko2010} and \citet{Liu2020}, whereas \citet{McClure2017} observed comparable accuracy between ODI and PPE) may be more closely linked to the specifics of the problem setup rather than inherent differences in the methods themselves. However, it is essential to recognize the inherent advantages of ODI. Unlike the PPE, ODI offers easier implementation of boundary conditions and handling of missing data within the domain (void regions). This flexibility makes ODI a valuable alternative, especially in scenarios where the management of complex geometries and incomplete data is critical. Furthermore, the ODI helps minimizing noise through regularization \citep{Pryce2024}.

The Green's function integral (GFI) method, recently proposed by \citet{Wang_2023}, utilizes the Green's function of the Laplacian operator to relate the instantaneous pressure field to the pressure gradient via a convolution kernel. Both the OS\-/MODI and GFI methods enable generalized implementation in two- and three-dimensional problems with arbitrary geometries, enhancing computational efficiency. However, both approaches are mesh-based, i.e., require connectivity of spatial elements, which can be particularly challenging when working with particle tracking data \citep{Neeteson2015}.

To address the complexities associated with mesh generation, this work proposes a novel, meshless approach that leverages automatic differentiation through parametric models to tackle the pressure integration task seamlessly. This method shares similarities with the recent work by \citet{Sperotto2022}, which utilized a meshless technique based on radial basis functions to derive pressure fields from scattered and noisy velocity data. However, our approach builds upon the concept of neural implicit representation to effectively integrate the pressure gradient, offering a straightforward yet efficient solution for pressure reconstruction. Given recent advancements that have established volumetric measurements as a new standard in PIV and LPT, it is essential for pressure reconstruction methods to effectively handle this type of data. Besides being meshless, our approach is capable of effectively handling void regions and noisy data, allowing it to work with complex volumetric datasets while enhancing overall effectiveness in pressure reconstruction.

\section{Methodology}
\label{sec:Methodology}

This section provides a concise overview of the proposed neural network approach for integrating Eq. \ref{eq:momentum} to evaluate the pressure field. Additional methods utilized in this study, including OS\-/MODI and GFI, are detailed in the Appendix.

\subsection{Sinusoidal Representation Network (SIREN)}
\label{sec:Methodology:SIREN}

We employ a neural network that parameterizes a function  $\Phi: \mathbf{x} \mapsto \Phi(\mathbf{x})$ to map spatial or spatiotemporal coordinates $\mathbf{x} \in \mathbb{R}^n$ to some quantity of interest (pressure, in this case) while satisfying a set of M constraints $\mathcal{C}_m$ on their domain $\Omega_m$: 
%
%
\begin{align}
    \mbox{find} \ \Phi(\mathbf{x}) \quad
    &\mbox{s.t.} \ \mathcal{C}_m \left( \nabla_\mathbf{x} \Phi(\mathbf{x}) \right) = 0 \\
    &\forall \mathbf{x} \in \Omega_m, \quad m = 1, \ldots, M \mbox{.}
    \nonumber
\end{align}

Here, we follow the SIREN model \citep{siren} and 
parameterize $\Phi$ as a fully connected neural network with the sine function as a periodic activation function:
%
\begin{align}
    \Phi(\mathbf{x}) &= \mathbf{W}_n ( \phi_{n-1} \circ \phi_{n-2} \circ \ldots \circ \phi_0 ) (\mathbf{x}) + \mathbf{b}_n,
    \nonumber
    \\
    \mathbf{x}_i &\mapsto \phi_i(\mathbf{x}_i) = \sin( \omega_i \mathbf{W}_i \mathbf{x}_i + \mathbf{b}_i ) \mbox{,}
    \label{eq:siren}
\end{align}
where $\phi_i : \mathbb{R}^{M_i} \mapsto \mathbb{R}^{N_i}$ is the $i^{th}$ layer of the network that applies on the input $\mathbf{x}_i \in \mathbb{R}^{M_i}$ an affine transformation defined by the weight matrix $\mathbf{W}_i \in \mathbb{R}^{N_i \times M_i}$ with a scaling factor $\omega_i \in \mathbb{R}$ and the biases $\mathbf{b}_i \in \mathbb{R}^{N_i}$, followed by a sine nonlinearity. The weights are initialized with $\text{w}_i \sim \mathcal{U}(-\sqrt{6/n}, \sqrt{6/n}) \mid i > 0$, being those from the first layer $\text{w}_0 \sim \mathcal{U}(-1/n, 1/n)$. Then, the resulting optimization problem is solved using gradient descent. The dataset $\mathcal{D} = \{ (\mathbf{x}_i, \ \nabla_\mathbf{x} \Phi(\mathbf{x}_i)) \}_i$ consists of a set of tuples of coordinates $\mathbf{x}_i = (x_i, \ y_i)$ within the domain $\Omega = \{\mathbf{x}_i \in \mathbb{R}^2 \}$, $\forall i \in \{1, \dots, N \}$, along with samples of $\nabla_\mathbf{x} \Phi(\mathbf{x})$ from the constrains $\mathcal{C}$.

The scaling factor $\omega_i$ in Equation \ref{eq:siren} is an empirical user-defined parameter chosen to stabilize the training process of SIREN and ensure its convergence. As highlighted by \citet{novello2024}, the frequencies generated by the network are primarily determined by the input layer, while the hidden layers control their amplitudes. To effectively manage the frequency bandlimit in SIREN, the input layer $\omega_0$ is set differently from the hidden layers. This distinct setting helps regulate the frequency characteristics more effectively during training. The selection of the scaling factor $\omega_i$ is detailed throughout the text.

To solve Eq. \ref{eq:momentum}, the representation $\Phi$ is sought in order to minimize the loss function:
%
\begin{equation}
    \mathcal{L} = \int_\Omega 
    \left\lVert
    \nabla_\mathbf{x} \Phi(\mathbf{x}) - \nabla p(\mathbf{x})
    \right\rVert
    \mbox{d} \mathbf{x}
    \mbox{,}
\end{equation}
where $\nabla p(\mathbf{x})$ is estimated from discrete samples of the material acceleration. Specifically, we make use of the mean squared error loss. While this loss is essentially equivalent to the cross-entropy loss associated with a unit Gaussian output distribution, which suggests that sharp signals would be attenuated, this does not occur here. The SIREN is capable of bypassing the spectral bias \citep{spectral_bias2019}, allowing it to effectively represent detailed signals \citep{tancik2020}. However, if the training process naturally introduces a denoising or smoothing effect, it could be particularly beneficial for solving inverse problems, especially when working with noisy experimental data. This attribute aligns with recent advancements in the ODI method, which we intend to leverage. In this context, spectral control is achieved through the network initialization.

The model was trained on a NVIDIA GeForce RTX 2070 GPU (8 GB of memory) utilizing the ADAM optimizer, with a learning rate set to $3 \times 10^{-5}$ over a span of $2,000$ epochs. This choice of epoch count was made to secure a robust approximation of the reconstructed pressure field. However, as evidenced by the results (not shown here for brevity), it is noteworthy that satisfactory reconstructions can be achieved with fewer epochs.

\subsection{Datasets}

In this work, we utilize the ``isotropic1024coarse'' dataset from the Johns Hopkins University Turbulence Database (JHTD) \citep{Li2008} to synthetically generate a planar velocity field from DNS solutions, as it provides ground truth pressure data. Noise is directly incorporated into the three-dimensional velocity field, and subsequent derivatives are evaluated, thereby circumventing errors associated with 2D flow approximations/assumptions \citep{McClure2017, deKat2012}. The material acceleration is computed using an Eulerian framework following the necessary operations described by \citet{Zigunov2022} to balance the velocity/pressure fields. Derivatives are taken using a second-order central finite difference scheme for both space and time. Subsequently, a two-dimensional slice is selected from the volumetric domain for analysis. The amplitude of the noise follows a normal distribution based on the maximum velocity amplitude within the slice, while the direction of velocity perturbation adheres to a uniform distribution over polar and azimuthal angles. In addition to utilizing the isotropic turbulence dataset from the JHTD dataset, we will incorporate experimental data from the bluff body wake of a cylinder with a slanted afterbody, as investigated by \citet{Zigunov2020_bluffbody}.

\section{Results}
\label{sec:results}

This section presents results obtained by the SIREN, OS\-/MODI and GFI for structures and unstructured meshes, as well as for numerical and experimental datasets of turbulent flows. Frames 39 to 41 of the JHTDB database 
are utilized to compute the temporal derivative, and subsequently, we extract the latest 100 data points along the $x$ and $y$ directions, and 3 points along the $z$ direction to form a subdomain. Noise with a uniform spherical distribution is introduced into the three-dimensional velocity field to evaluate the source term. This selection results in a mesh comprising 10,000 points after constraining the analysis to a 2D slice (the second $x$-$y$ plane). This implies that only the $x$ and $y$ components of the material acceleration are fitted by the network.

\subsection{Uniform structured mesh}
\label{sec:results-uniform}

\begin{figure*}[!htb]
    \centering
    \includegraphics[width=1\textwidth]{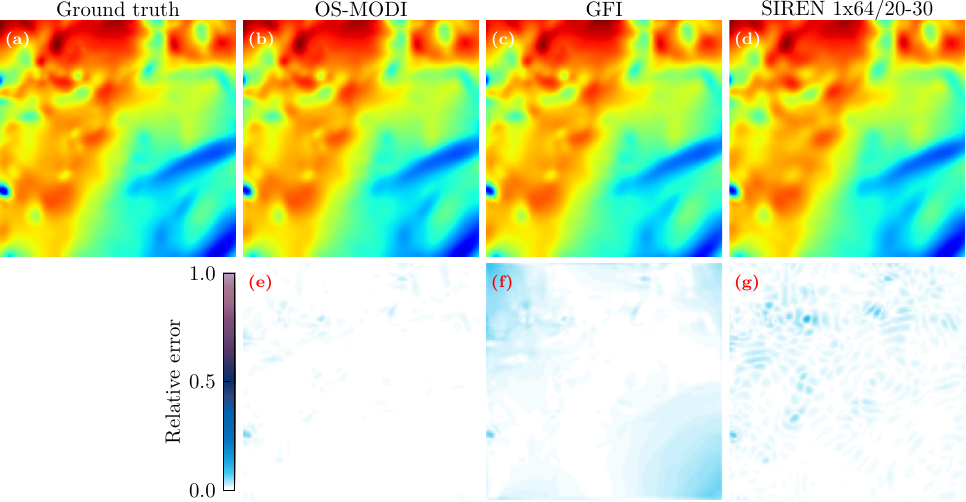}
    \caption{Pressure reconstruction with no noise added to the velocity field (top row) and relative error map (bottom row).}
    \label{fig:error_0_noise}
\end{figure*}

After computing the material derivatives, we proceed to assess the accuracy of the reconstructed pressure field using various solvers. The comparison of results obtained with different solvers under zero noise conditions is presented in Fig. \ref{fig:error_0_noise}. From this figure, it is evident that the solutions converge closely to the ground truth, as depicted in the top row. However, upon closer examination of the bottom row, it becomes apparent that all methods exhibit a noticeable error. This behavior is likely attributed to the absence of out-of-plane information related to the pressure derivative, coupled with the potential propagation of this error from the boundaries. Such an issue has been previously discussed in the works of \citet{Pan2016}, \citet{Faiella2021}, and \citet{Nie2022}. For this analysis, we present only the SIREN results for the 1x64/20-30 architecture, which features a single hidden layer with 64 nodes. The numbers following the slash, $\omega_0 = 20$ and $\omega_{i>0} = 30$, correspond to the input frequency scaling and the hidden layer scaling, respectively. This particular architecture is chosen for its balance in environments with varying levels of noise. The impact of the network architecture and scaling factors will be discussed later in the text.

\begin{figure*}[!htb]
    \centering
    \includegraphics[width=1\textwidth]{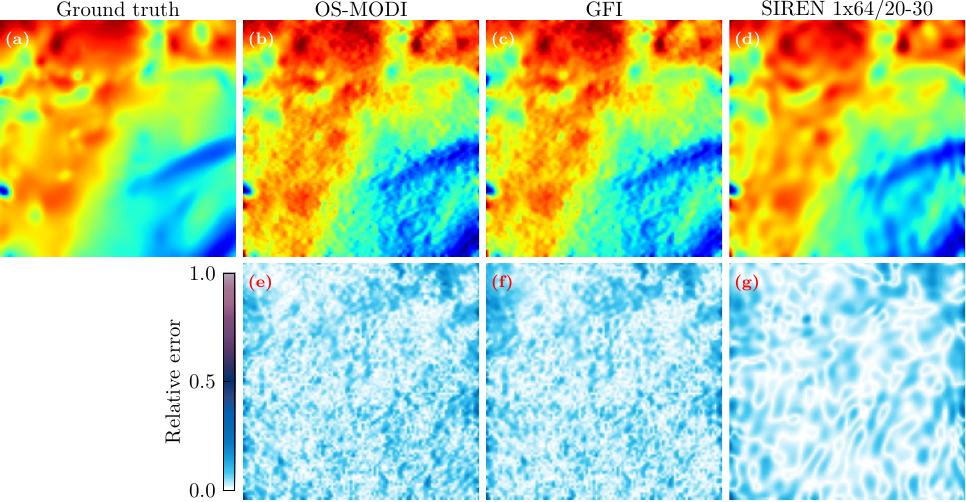}
    \caption{Pressure reconstruction with 10\% noise added to the velocity field (top row) and relative error map (bottom row).}
    \label{fig:error_10_noise}
\end{figure*}

The results for the worst-case scenario with 10\% added noise are depicted in Fig. \ref{fig:error_10_noise}. From this figure, it is evident that the presence of erroneous data significantly compromises the accuracy of the reconstructed pressure field. Notably, the results obtained from the OS\-/MODI and GFI techniques exhibit a striking similarity, rendering them almost indistinguishable. In contrast, the SIREN method produces a smoother solution, albeit following similar trends to the other techniques. Despite the challenges posed by the corrupted data, 
all methods demonstrate satisfactory performance in pressure reconstruction, highlighting their robustness under adverse conditions.

\begin{figure}[!htb]
    \centering
    \includegraphics[width=1\columnwidth]{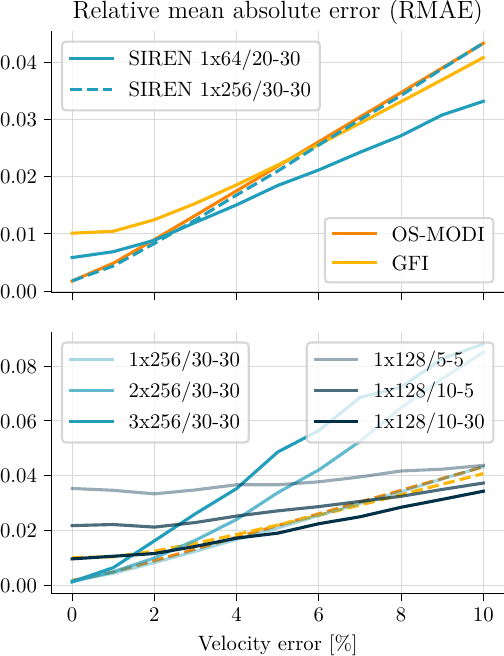}
    \caption{Relative MAE for different levels of velocity error (based on a $\pm 2 \sigma$ standard deviation).}
    \label{fig:graph_error_10k}
\end{figure}

The error assessment is provided in Fig. \ref{fig:graph_error_10k}, where the evaluation is conducted relative to the maximum amplitude of the ground truth signal. From the results shown in the top graph, it is clear that both the OS\-/MODI method and the SIREN 1x256/30-30 architecture - featuring a single hidden layer with 256 nodes, and input and hidden layer frequency scalings of $\omega_0 = 30$ and $\omega_{i>0} = 30$, respectively - are virtually indistinguishable and exhibit minimal error when no noise is added to the velocity field. However, as the level of noise increases, the performances of all methods become comparable, scaling almost linearly with increasing velocity error. However, the GFI solution, in the zero-noise case, shows a pronounced mean absolute error (MAE). This observation aligns with our earlier discussion regarding the GFI method displaying a more pronounced MAE under zero noise conditions. 
%

As the level of added noise rises, the SIREN method using the 1x64/20-30 architecture emerges as the most effective among the evaluated techniques (see the top plot in Fig. \ref{fig:graph_error_10k}). By varying the architecture — specifically, by selecting the number of hidden layers, nodes and scaling factors — one can control the model's response to noisy data. For instance, choosing a proper combination of scaling factors in a network with fewer layers and nodes can filter out high-frequency information, enhancing effectiveness in handling noise.

As demonstrated in the results, even a neural network architecture with a single hidden layer consisting of 64 nodes can effectively represent complex signals. The theoretical and experimental work by \citet{novello2024} provides insights into the smoothness and high representation capacity of the SIREN model. According to the authors, the SIREN is capable of capturing detailed signals due to the exponentially growing number of frequencies generated by the sine activation function. Specifically, the first layer of the SIREN serves as a Fourier feature mapping \citep{benbarka2021}, projecting the input coordinates onto a set of sines and cosines. Each sinusoidal neuron in the hidden layer expands into a broad sum of sines, with the new frequencies being integer linear combinations of the input frequencies (i.e., the weights of the input layer). The corresponding amplitudes of these sinusoidal components are determined by the hidden weights and exhibit exponential decay as the integer coefficients of the frequency compositions increase. As a result, the amplitudes in the expansion of each sinusoidal neuron remain bounded. This boundedness, coupled with the initialization process, generates frequencies that progressively surround the input frequencies, enhancing the network's capacity to represent intricate details.

Building on the spectral representation mechanism of the SIREN, we control the bandlimit at initialization by defining both the network architecture and scaling factors. Since the initial weights are sampled from $\text{w}_0 \sim \mathcal{U}(-1/n, 1/n)$, they do not introduce high frequencies at the beginning of the training. To properly scale the initial frequencies for the problem at hand, we introduce the scaling factor $\omega_0$. Specifically, we consider the smallest wavenumber $k_{\min} = 2 \pi / \max(L_x, L_y)$ present in the signal, where $L_x$ and $L_y$ represent the spatial extents of the domain along the $x$ and $y$ directions, respectively. The initial frequency scale is then set to $\omega_0 = c \, k_{\min}$, where $c$ is a constant. In our experiments, we found that values of $2 \lessapprox c \lessapprox 3$ yielded good results across various test cases\footnote{For instance, in our previous results, the domain corresponds to the last 100 points of the JHTDB ``isotropic1024coarse'' dataset, where $L_x = L_y \approx 0.6136$ and $k_{\min} = 10.24$. We set $\omega_0 = 20$, which implies $c \approx 2$.}.

According to \citet{novello2024}, the hidden layers determine only the frequency amplitudes, with the hidden neurons sharing the same harmonics. Therefore, $\omega_i$, for $i > 0$ primarily affects the rate at which the amplitudes of the frequencies grow through backpropagation during training. However, it is important to note that setting large values for $\omega_i$ can lead to instabilities, particularly due to the risk of the exploding gradient problem. Our experiments suggest that setting $\omega_i$ in the range $20 \lessapprox \omega_i \lessapprox 30$ results in favorable performance.

Since the new frequencies generated by the sinusoidal neurons are integer linear combinations of the input frequencies (weights of the input layer), and their amplitudes are bounded by a small integer \citep{novello2024}, increasing the number of neurons in the hidden layer enhances the expressive capacity of the Fourier feature mapping, enabling the representation of a broader range of frequencies. This is particularly important when representing pressure signals with a wide range of scales. As shown in the top plot of Fig. \ref{fig:graph_error_10k}, increasing the number of neurons from 64 to 256 made the results from SIREN almost indistinguishable from those of OS\-/MODI. The expanded dictionary of input frequencies allowed for a richer representation, leading to a more accurate solution in noise-free conditions, while being more sensitive to noise in challenging environments.

Increasing the number of hidden layers also contributes to expanding the resolution capacity of the network. The logic behind this is similar to that of a single hidden layer: the new frequencies learned by each hidden layer act as the input frequencies for the subsequent layer. Thus, the composition of additional hidden layers follows the same principles, with the key difference being that the input frequencies are now those learned by the previous layer. This approach, however, is more prone to learning higher frequencies. This is illustrated in the bottom plot of Fig. \ref{fig:graph_error_10k} (blue lines), which shows that as more hidden layers are added, the model becomes more sensitive to noise, leading to an increase in error as the noise level rises.

Figure \ref{fig:graph_error_10k} also illustrates how improper initialization of parameters can negatively impact pressure evaluation, even in architectures with more nodes, such as 128 (see gray lines). When the largest input frequency, which is related to $\omega_0$, is set too low relative to the problem’s requirements, and $\omega_{i>0}$ is too small, the model ends up prioritizing the learning of only low frequencies. This explains why the error of the 1x128/5-5 model remains almost constant across the entire noise range. Increasing $\omega_0$ to 10 improves the result, but maintaining a low $\omega_{i>0}$ (such as 5 in this case) slows down convergence.

\begin{figure}[!htb]
    \centering
    \includegraphics[width=1\columnwidth]{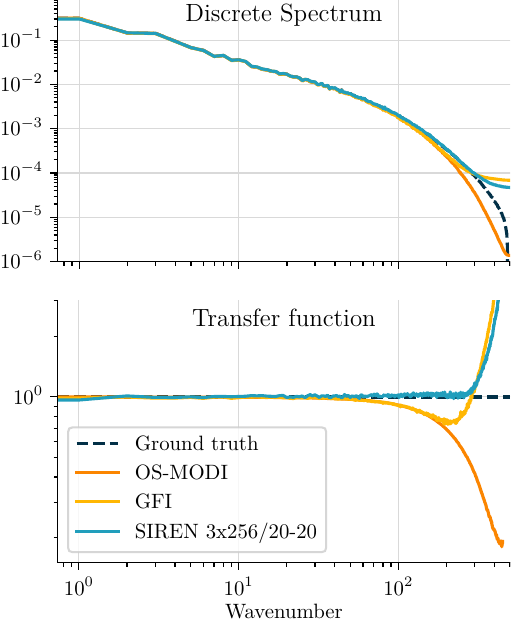}
    \caption{Spectral analysis of different pressure reconstruction methods from a 2D slice of the noiseless JHTDB ``isotropic1024coarse'' dataset.}
    \label{fig:spectral_all}
\end{figure}

Figure \ref{fig:spectral_all} presents the spectral analysis of various pressure reconstruction methods applied to a 2D slice of the noiseless JHTDB ``isotropic1024coarse'' dataset. The analysis includes the discrete spectrum and the transfer function (numerical reconstruction divided by the ground truth) for the 40th snapshot and the first 1024x1024 z-plane slice. Given that the domain is periodic, a spectral derivative is utilized to evaluate the ground truth spectrum. From Fig. \ref{fig:spectral_all}, it is evident that both OS\-/MODI and GFI methods accurately resolve up to a wavenumber of 30. Beyond this point, the solutions encounter a noise floor that introduces artifacts and excites multiple frequencies. In contrast, the SIREN 3x256/20-20 provides a more accurate solution, only starting to fail at a wavenumber approximately one order of magnitude higher. While OS\-/MODI and GFI struggle well before this point, the SIREN maintains high accuracy across a much broader range of frequencies. The limitation at higher wavenumbers is likely due to the use of single precision, where machine errors begin to affect the results at this level of resolution. Additionally, the use of an MSE loss, which is equivalent to negative log-likelihood with Gaussian priors, may also contribute to this limitation.
%

\begin{figure}[!htb]
    \centering
    \includegraphics[width=1\columnwidth]{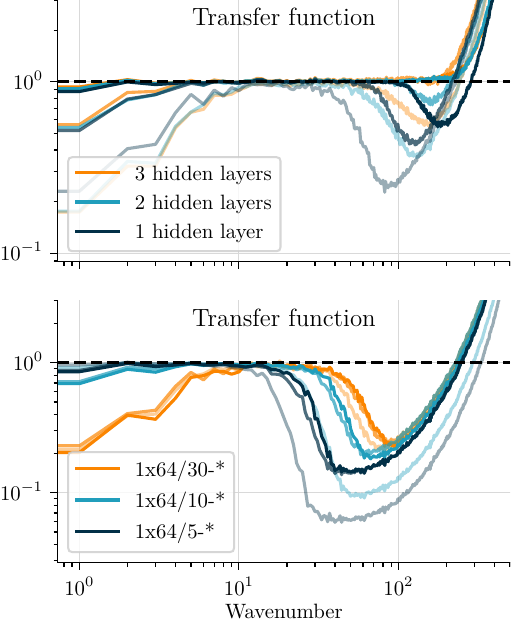}
    \caption{Spectral analysis of different SIREN architectures from a 2D slice of the noiseless JHTDB ``isotropic1024coarse'' dataset. In the top plot, curves are color-coded for the number of nodes in each hidden layer: lighter, intermediate, and darker shades represent 64, 128 and 256 nodes, respectively. In the bottom plot, curves are color-coded for the scaling of the hidden layer: lighter, intermediate, and darker shades represent $* = \omega_{i>0} = 10$, 20 and 30, respectively.}
    \label{fig:spectral_sirens}
\end{figure}

In Fig. \ref{fig:spectral_sirens}, we explore how the choice of architecture and scaling factors influences the frequency response of the results, highlighting the varying capabilities of different configurations in capturing spectral information. Specifically, we trained networks with varying numbers of hidden layers, nodes per hidden layer, and scaling factors. Configurations tested include 1, 2, 3 and 4 hidden layers, as well as 64, 128, and 256 nodes per layer. Additionally, we explored scaling factors $\omega_i$ with values of 2, 5, 10, 20, 30 and 40. Some of these results are shown in Fig. \ref{fig:spectral_sirens}. In the top plot of this figure, we present results for varying the number of layers and nodes, with fixed scaling factors of $\omega_0 = 30$ and $\omega_i = 20$ for $i > 0$, corresponding to architectures labeled, for example, as 1x256/30-20. The transparency levels represent the number of nodes, with 64 nodes shown as the most transparent and 256 nodes as the least transparent. As demonstrated in the top plot, increasing the number of nodes and hidden layers allows the neural network to resolve higher frequency content more effectively.
%

When the number of nodes is too low to capture the problem's bandlimit, as seen with 64 nodes, both the lower and higher frequencies are not resolved adequately with the given initialization of $\omega_0$. This issue is addressed by increasing the number of nodes, which leads to a richer initial dictionary of input frequencies. As the network trains, new frequencies emerge from combinations of these initial frequencies, and with more nodes, the network has the capacity to generate a broader range of frequencies. This expanded frequency representation enables the network to better capture both low and high-frequency components, improving the overall resolution of the frequency spectrum. Increasing the number of nodes primarily enhances the resolution of lower frequencies, as it allows for more low-frequency components to be combined and form new low frequencies. In contrast, increasing the number of hidden layers facilitates the combination of these frequencies into new ones, with a greater likelihood of emphasizing high-frequency content. This relationship helps explain the performance of the deeper networks in Fig. \ref{fig:graph_error_10k} under noisy environments.

The bottom plot of Fig. \ref{fig:spectral_sirens} illustrates the effect of the scaling factors. A simple architecture with a single layer of 64 nodes is used as an example, corresponding to the same parameters employed in the previous results presented in this work. In this plot, the transparency levels indicate the scaling factor of the hidden layers, with values of 10 (most transparent), 20, and 30 (least transparent). From the figure, we observe that initializing the network with $\omega_0 = 5$ prioritizes lower frequencies, as expected. However, increasing $\omega_0$ enables the network to capture higher frequencies, albeit at the cost of compromising lower frequencies, particularly when the dictionary of input frequencies is insufficient, as seen with 64 nodes. Increasing the scaling factor $\omega_i$ of the hidden layers predominantly affects the learning of higher frequencies, with minimal impact on the representation of lower frequencies.

The preceding solutions are obtained using a regular mesh, a common practice in PIV experiments. These results show the efficacy of all three methods in accurately reconstructing pressure fields from PIV data. Regarding computational efficiency, particularly in the present 2D spectral analyses, the GFI method incurred the highest computational expense, taking approximately 201 minutes to run. Conversely, the OS\-/MODI method exhibited the most cost-effective computational performance, taking only 2 minutes\footnote{It is worth noting that we did not use the optimized implementation of the OS-MODI, as proposed in \citep{zigunov2024oneshot}.}, followed by the SIREN 3x256/20-20, which took 16 minutes (training for 2,000 epochs)\footnote{Other architectures of the SIREN network exhibit different execution times; for instance, the 2x256 architecture takes approximately 12 minutes, while the 3x128 architecture requires about 6 minutes.}. While the GFI method entails solving a smaller linear system (confined to boundary elements), computational bottlenecking arises primarily from the evaluation of the volume integral. To improve the computation speed of the GFI method, Python functions were translated into optimized machine code at runtime using Numba v0.60 \citep{numba}. This enabled significant performance enhancements, particularly by parallelizing loops and leveraging Just-in-Time (JIT) compilation. The method was executed on a machine equipped with an Intel\textsuperscript{\textregistered} Core\textsuperscript{TM} i7-7700 CPU running at 3.60GHz with 8 cores, ensuring multi-core processing to maximize computational throughput. However, it is important to note that neither method was optimized for computational performance, as we relied on traditional matrix inversion libraries available in Python. As such, a definitive comparison of the computational performance of the methods cannot be provided at this point, with such task being beyond the scope of this study.

\subsection{Unstructured meshes}
\label{sec:results-unstructured}

Evaluating the performance of the methods, particularly the SIREN, in an unstructured mesh setting is imperative due to its widespread application in contexts involving complex geometries or LPT data. In this regard, adherence to maximum principles is essential for ensuring the reliability and accuracy of numerical methods. However, unlike the Cartesian mesh used for PIV data, discrete maximum principles do not hold for approximations constructed on arbitrary meshes \citep{Korotov2001}.

To assess the influence of mesh quality on our solutions, we focus exclusively on scenarios with zero noise introduced to the velocity field. In this context, we bypass the need for calculating the material derivative from numerical velocity data, opting instead to directly compute the source term by deriving the pressure field gradient. This approach minimizes potential numerical errors inherent in the direct evaluation of the material derivative.

\begin{figure*}[!htb]
    \centering
    \includegraphics[width=1\textwidth]{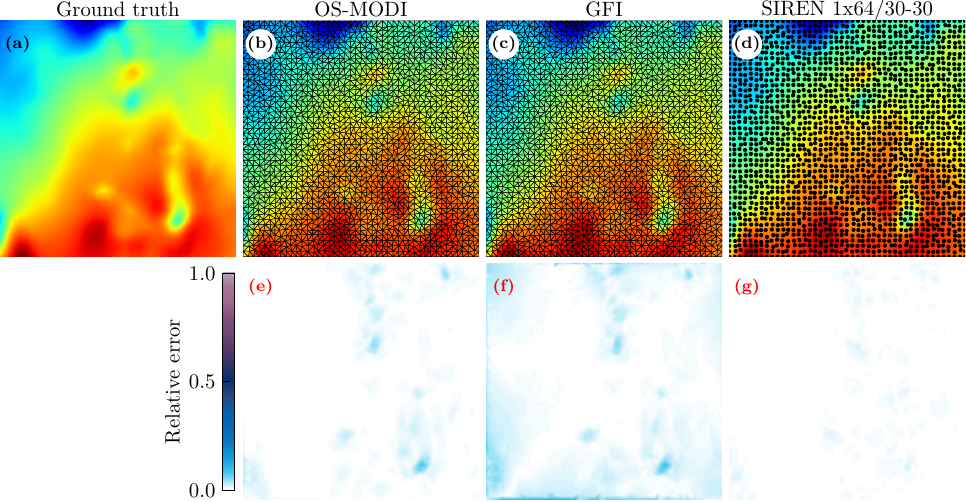}
    \caption{Assessment of the pressure reconstruction on a well-conditioned  unstructured triangular mesh.}
    \label{fig:unstructured_nice_boundaries}
\end{figure*}

\begin{figure*}[!htb]
    \centering
    \includegraphics[width=1\textwidth]{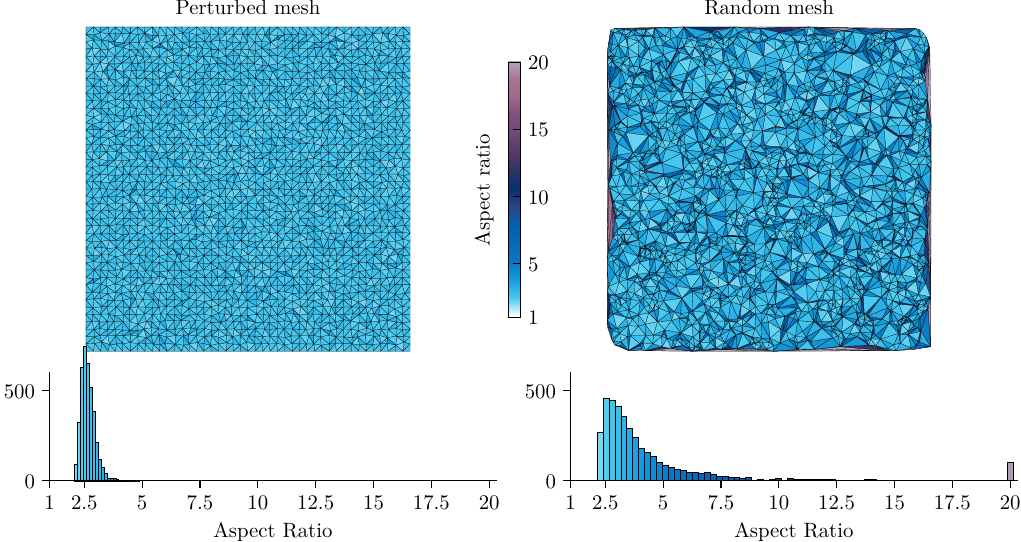}
    \caption{Comparison of triangle aspect ratios in Delaunay meshes: uniformly-perturbed seed vs. random seed}
    \label{fig:mesh_quality}
\end{figure*}

We start with a meticulously arranged set comprising nearly 2,000 points, initially evenly distributed across a 2D Cartesian grid. To explore the solution at a different time, we depart from the selection of the 40th frame of the JHTDB ``isotropic1024coarse'' database and instead utilize the 1st frame. For this analysis, we select a 100x100 subdomain from the 1024x1024 grid, starting at the indices (400, 900). Subsequently, we introduce slight perturbations to the coordinates of internal vertices, following $\textbf{x}_i = (x_i + \varepsilon, y_i +\varepsilon)$ where $ \varepsilon \sim \mathcal{N}(0, (4\mathrm{e}{-4})^2)$, such that $\textbf{x}_i$ is not on the boundary $\partial \Omega$. Employing Delaunay triangulation on this seed of points, we construct the mesh utilized in the OS\-/MODI and GFI solvers. The resultant mesh is depicted in Figs. \ref{fig:unstructured_nice_boundaries} (b) and (c), as well as Fig. \ref{fig:mesh_quality} (labeled ``Perturbed mesh'').

Overall, the pressure reconstructions from all methodologies have good agreement with the ground truth (see Fig. \ref{fig:unstructured_nice_boundaries}). The SIREN 1x64/30-30 exhibits superior performance, boasting a MAE of 0.0019, compared to 0.0029 for the OS\-/MODI and 0.0081 for the GFI. It is important to note that aliasing errors are introduced when resampling the original 10,000 grid points into 2,000. To mitigate the influence of these aliasing errors on the assessment of the methods, the ground truth solution was also interpolated to the same 2,000 grid points, and the reconstructed pressure fields were compared against this interpolated ground truth.

The use of well-distributed points tend to lead to well-conditioned meshes with favorable geometric properties, which are more likely to preserve maximum principles \citep{miller1995}. However, the non-uniform distribution of LPT data \citep{Neeteson2015, Neeteson2016} results in variations in mesh density across the domain. In regions with sparse mesh density, the numerical solution may not accurately capture the behavior of the underlying physical problem. Furthermore, clustered or unevenly distributed points result in a mesh containing elements with poor quality. These combined factors can compromise the adherence to maximum principles.

\begin{figure*}[!htb]
    \centering
    \includegraphics[width=1\textwidth]{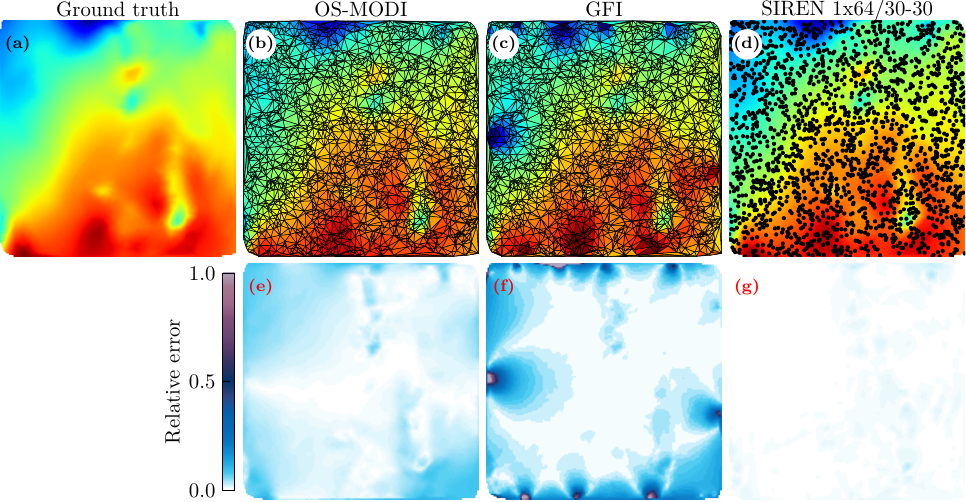}
    \caption{Assessment of the pressure reconstruction on a ill\-/conditioned unstructured triangular mesh with uniformly distributed seed of points.}
    \label{fig:unstructured_bad_boundaries}
\end{figure*}

To demonstrate the influence of different point seeding on the solution, we revisit the same frame as before but opt to resample the 2,000 points. Here, we sample the points such that $(x_i, y_i) \sim \mathcal{U}(-L/2, L/2)$, where $L$ represents the size of the domain (100 grid points in both the $x$ and $y$ directions starting at the indices (400, 900)) and $\mathcal{U}$ is the uniform random distribution. Employing Delaunay triangulation on this set of points yields the mesh depicted in Figs. \ref{fig:unstructured_bad_boundaries} (b) and (c), as well as in Fig. \ref{fig:mesh_quality} (labeled ``Random mesh''). 

Utilizing the solvers on the unstructured mesh generated from this seed of uniformly distributed points, we produce the reconstructions showcased in Fig. \ref{fig:unstructured_bad_boundaries}. Notably, in this instance, the SIREN method outperforms the others, exhibiting a relative MAE of 0.0031, compared to 0.0213 for OS\-/MODI and 0.0401 for GFI. The superior performance of the SIREN method is evident in the considerably lower error visualized in Fig. \ref{fig:unstructured_bad_boundaries} (g). Conversely, the GFI reconstruction exhibits pronounced errors on the domain boundaries (refer to Figs. \ref{fig:unstructured_bad_boundaries} (c) and (f)).

To gain insight into this behavior near the boundary, we analyze the aspect ratios of the triangles obtained from the uniform seed of points in Fig. \ref{fig:mesh_quality} and compare them with the well-conditioned mesh used in Fig. \ref{fig:unstructured_nice_boundaries}. The aspect ratio, defined as the ratio between the circumradius and the inradius of the triangle, serves as a robust and easily computed metric for assessing the quality of generated meshes. In Fig. \ref{fig:mesh_quality} we present histograms of the aspect ratios for each mesh, with a maximum aspect ratio saturated to 20 for visualization purposes only, although some triangles achieve aspect ratios much higher than this value.
%

The aspect ratio provides a measure of how close the triangle is to being equilateral ( aspect ratio of 1). Small angles in mesh triangles can lead to highly elongated or degenerate triangles, causing numerical instability and ill\-/conditioning in finite volume methods, especially when solving equations involving gradients or derivatives like the Poisson equation. Ill\-/conditioned systems result from small angles amplifying errors in the numerical solution, leading to inaccuracies and instability. Additionally, small angles can induce significant coefficient variations in discretized equations, heightening sensitivity to numerical errors and round-off effects. The presence of skewed triangles is influenced by the arrangement and density of points, as well as the assumption of a convex hull for the domain.

Upon comparing the aspect ratios of the triangles depicted in Fig. \ref{fig:mesh_quality} with the results obtained from the GFI method in Fig. \ref{fig:unstructured_bad_boundaries} (c) and (f),
it becomes evident that the errors stem from ill\-/conditioned triangles. This issue arises as a consequence of employing the constant midpoint rule to evaluate the integrals of the boundary elements in our implementation. The close proximity of source points (i.e., the centroids of the triangles) to the singularities on the boundary elements causes the solver to produce inaccurate results. These inaccuracies propagate inside the domain and contaminate the solution when solving the Poisson equation, where accuracy near boundaries is critical \citep{Neeteson2016, Pan2016, Nie2022}. Despite the error being smaller in the OS\-/MODI solver, as it does not need to treat singular integrals, we also observe a significant contamination of the error inside the domain. In future work, we plan to employ more robust schemes to mitigate this issue with the singularity \cite{wolf2011}. Nevertheless, we present these results to highlight potential challenges that may arise when applying the method to a na\"ively-generated mesh.

\subsection{Bluff body wake experiment}

To demonstrate the effectiveness of SIREN in analyzing real PIV data, we examine the flow over a slanted bluff body, as studied by \citet{Zigunov2020_bluffbody}. This configuration consists of a cylinder inclined at a 45-degree angle, operating at a diameter-based Reynolds number of Re = 25,000. At this Reynolds number, flow separates at the leading edge of the slanted surface, reattaching downstream and forming a separation bubble, along with a counter-rotating vortex pair. To capture the dynamics of this bubble, a planar PIV experiment was conducted in the streamwise direction, collecting 500 fields at a rate of 10 Hz. A LaVision sCMOS camera, paired with a 200 mJ per pulse Quantel Evergreen laser, was utilized for image acquisition. Details are provided in \citet{Zigunov2020_bluffbody}.
\begin{figure*}[!htb]
    \centering
    \includegraphics[width=1\textwidth]{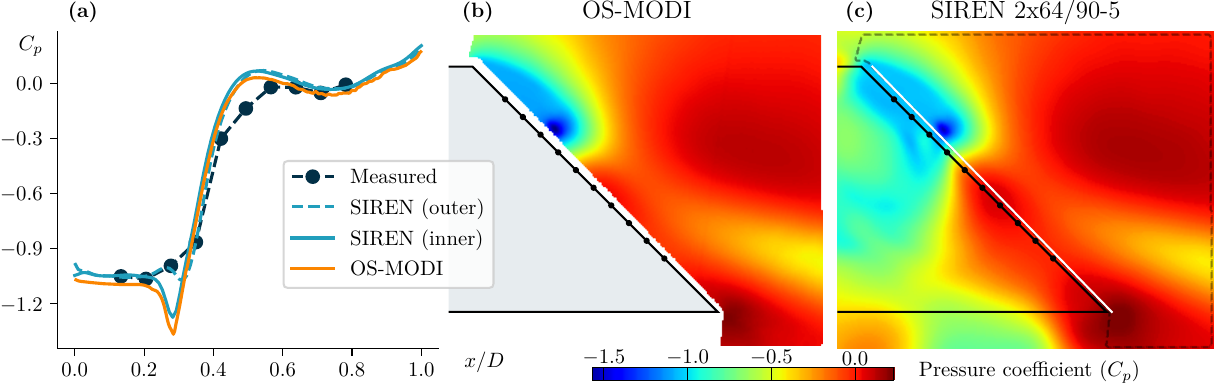}
    \caption{(a) Comparison of the proposed SIREN method across two distinct regions (inner and outer) with OS\-/MODI results and conventional surface measurements from traditional transducers. Contours of center-plane pressure field obtained from PIV measurements for the slanted cylinder model using the (b) OS\-/MODI method and (c) the proposed SIREN method.}
    \label{fig:bluff_body}
\end{figure*}

The results obtained using the SIREN 2x64/90-5\footnote{For the bluff body, $L = \max(L_x, L_y) = 0.2243$, which corresponds to a wavenumber $k = 2 \pi / L \approx 28.6$. By setting $c \approx 3$, we obtain $\omega_0 = c \, k = 90$.} method for this dataset are shown in Fig. \ref{fig:bluff_body}, alongside those from OS\-/MODI and Omega PX653 pressure transducers located at the slanted wall of the bluff body. SIREN, as a continuous implicit representation, can interpolate and extrapolate the solution. This enables two distinct solutions: the inner solution, which corresponds to the valid fluid region, and the outer solution, which extends beyond it. The inner solution is derived along the white line in subplot (c), while the outer solution is extrapolated along the slanted surface of the bluff body. Subplot (a) compares the pressure estimates from OS\-/MODI and SIREN, showing that both methods tend to underestimate pressure at the vortex core due to the white line offset, caused by laser reflection. While OS\-/MODI and SIREN (inner and outer) results are similar, the SIREN (outer) solution, extrapolated along the slanted surface, should not be relied upon. Extrapolated values can vary significantly and may yield non-physical results, depending on the neural network architecture and the field topology; they are presented here for comparison purposes only. Subplots (b) and (c) display the pressure fields from OS\-/MODI and SIREN, respectively. Both show the bluff body centerline, with pressure transducer locations marked along the slanted surface. Subplot (c) also highlights the valid fluid region with a dashed black line, while the masked region appears blank in the OS\-/MODI solution (b).

\section{Conclusions}
\label{sec:conclusions}

A numerical approach is presented to accurately reconstruct pressure fields from image velocimetry data using a sinusoidal representation network. The proposed framework is shown to be more accurate than available techniques under noisy environments with proper tuning. Moreover, it works as a meshless technique, circumventing errors introduced by ill\-/conditioned meshes that may be generated from PIV/LPT data in complex flow configurations. 

First, the performance of various solvers is analyzed for reconstructing the pressure field on a regular Cartesian mesh, including the recently developed OS\-/MODI and GFI techniques, as well as the proposed SIREN approach. Results indicate that while both SIREN and OS\-/MODI methods outperform the GFI in noise-free conditions, the SIREN method exhibits superior performance in the presence of added noise. This is mostly due to the compactness property of sinusoidal networks, which is explained by the frequency factoring mechanism. Under adverse conditions, all methods demonstrate satisfactory accuracy, with the SIREN offering a smoother solution, which can be attributed to its flexibility in specifying bandlimit control at initialization, allowing for more effective filtering of the solution. In this context, we show that adding more layers and nodes in the SIREN enables the network to resolve higher spectral content. Increasing $\omega_0$ allows the network to capture higher frequencies, but at the cost of deteriorating the lower frequencies, especially when the number of nodes is low. Meanwhile, increasing the scaling factor $\omega_i$ of the hidden layers primarily influences the learning of higher frequencies, with little effect on lower frequencies. A network with fewer layers and nodes, however, is better at filtering out noise.

Transitioning to unstructured mesh settings, crucial for addressing complex geometries and LPT data, an examination of mesh quality reveals challenges associated with ill\-/conditioned triangles, particularly near boundaries. Meshes with large aspect ratio triangles can exacerbate numerical errors and introduce spurious oscillations in the solution. The presence of skewed triangles can be influenced by various factors, including the arrangement and density of points, as well as the assumption of a convex hull for the domain. Since completely eliminating them may be challenging, especially in complex or non-uniform point distributions, mesh quality considerations must be meticulously addressed. In regions where these triangles prevail, both the OS\-/MODI and GFI methods may yield inaccurate results, resulting in errors in the computed pressure field. This becomes particularly problematic near boundaries, where Neumann boundary conditions are applied. 

In the case of OS\-/MODI, the presence of ill\-/conditioned triangles can impede the convergence of the linear system solution, potentially leading to computational inefficiencies. Consequently, the numerical advantage typically associated with OS\-/MODI may be compromised. For the GFI method, our analysis highlights the need for careful attention to the computation of boundary integrals. The possibility of encountering ill\-/conditioned triangles near boundaries, arising from factors such as point distribution or the shape of the convex hull, underscores the potential for issues with singular integrals when using the constant midpoint rule. This emphasizes the importance of employing accurate schemes to effectively handle singularities and mitigate inaccuracies in the computation of boundary integrals, ensuring the stability and accuracy of the GFI method, particularly in regions where boundary effects are prominent.

In this context, the proposed SIREN method emerges as an appealing solution. Results show that it sidesteps the complexities associated with mesh generation, leading to lower relative errors along the entire fluid domain compared to both the OS\-/MODI and GFI. The SIREN can also inherently apply a denoising or smoothing effect to the inverse problem, which is stronger for shallower networks and lower values of $\omega_i$. This attribute proves advantageous when handling noisy experimental data, enhancing the overall efficacy of pressure field reconstruction.

\begin{acknowledgements}
The authors gratefully acknowledge Prof. Mark Glauser for facilitating Dr. Renato Miotto's visit to Syracuse University, where a substantial portion of this work was carried out. Additionally, the authors wish to express their sincere gratitude to Dr. Tiago Novello for his insightful discussions on the theory behind SIREN.
\end{acknowledgements}

\section*{Funding}

The authors acknowledge Fundação de Amparo à Pesquisa do Estado de São Paulo, FAPESP, for supporting the present work under research grants No. 2013/08293-7 and 2021/06448-0. FAPESP is also acknowledged for the fellowships provided to the first author under grants 2022/09196-4 and 2023/13431-1. Conselho Nacional de Desenvolvimento Científico e Tecnológico, CNPq, is also acknowledged for supporting this research under Grant No. 308017/2021-8. Finally, the authors acknowledge the Coaraci Supercomputer for computer time (FAPESP grant No. 2019/17874-0) and the Center for Computing in Engineering and Sciences at Unicamp (FAPESP grant No. 2013/08293-4).

\bibliographystyle{plainnat}
\bibliography{bibliography}   

\section*{Appendix}

In what follows, the solvers employed for integrating Eq. \ref{eq:momentum} with the synthetically generated data are outlined. These include the OS\-/MODI and the GFI methods. Both methods were validated by comparing their results to the analytical solution for the Taylor-Green vortex, as presented in \citet{Charonko2010}. The identical outcomes obtained in this comparison confirm the accuracy and reliability of the implementations.

\subsection*{One-Shot Matrix Omni-Directional Integration (OS\-/MODI)}
\label{sec:Methodology:OS-MODI}

Consider the conservative vector field $\nabla p = \mathbf{f}$ defined in a domain $\Omega$ containing a set of $N$ discrete (noisy) sources $\mathbf{f}(\textbf{x}_i), \ i=1, \dots, N$ that correspond to measurements of the right hand side of Eq. \ref{eq:momentum}. Following \citet{zigunov2024oneshot}, the pressure $p_c$ at a given mesh cell $c \in \Omega$ with neighboring cells $j$ can be calculated using the ``face-crossing'' scheme by solving the system
%
\begin{align}
    &p_c \sum_j \frac{\tilde{b}_j A_j}{A_{tot}} - \sum_j \frac{\tilde{b}_j A_j}{A_{tot}} p_j = \notag \\
    &- \sum_j \Delta j \frac{\tilde{b}_j A_j}{A_{tot}} \left( \frac{f_j(\textbf{x}_j) + f_j(\textbf{x}_c)}{2} \right) \quad \mbox{,} 
    \label{eq:os-MODI}
\end{align}
where $A_j$ is the corresponding shared face area with the neighboring point $\textbf{x}_j$ and $A_{tot} = \sum_j (b_j + \tilde{b}_j) A_j$ is the total area of the cell $c$. Here, $b_j$ is a boolean value for each cell, defining whether there is missing data (NaN) for the source field $\mathbf{f}$ in its neighboring face $j$, i.e.
\begin{equation}
    b_j = 
    \begin{cases}
        0, & \text{if $\textbf{f}(\textbf{x}_j) \neq$ NaN}\\
        1, & \text{if $\textbf{f}(\textbf{x}_j) =$ NaN,}
    \end{cases}
\end{equation}
and $\tilde{b}_j$ is the NOT boolean operation on $b_j$, to define where data is available instead. So, for example, if the face $j$ of a given cell $c$ faces a void region or the edge of the domain, there will be no point $\textbf{x}_j$ nor measurements $\textbf{f}(\textbf{x}_j)$, and $b_j = 1$ (or equivalently $\tilde{b}_j = 0$). The term $\Delta j$, in turn, stands for the vector from the centroids of cells $c$ to that of cell $j$ (\underline{not} the distance between the measurement locations $(\mathbf{x}_j - \mathbf{x}_c)$). Finally, the expression within the parenthesis represents the line integral of the pressure gradient, approximated using the trapezoidal rule. Here, $f_j$ denotes the component of the source vector $f_j$ along the direction connecting cells $c$ and $j$. It is worth mentioning that while alternative numerical schemes could be utilized for computing this integral, implementing them becomes challenging, particularly for unstructured meshes.

The resulting linear system from Eq. \ref{eq:os-MODI} is singular and needs regularization to make the problem well-posed. According to \citet{Pryce2024}, this regularization is naturally achieved through the solution of the system using the conjugate gradient (CG) method, which brings a key advantage beyond numerical considerations: it minimizes energy in the solution, thereby suppressing non-smooth components in the resulting pressure field. This phenomenon is complemented by the low-pass filter effect induced by the Laplacian, which is independent of the numerical solver utilized \citep{Faiella2021}. This Laplacian arises from the fact that solving Eq. \ref{eq:os-MODI} implies addressing a Poisson problem with a Neumann boundary condition, as elucidated by \citet{Pryce2024}. In addition, the regularization addresses the compatibility condition issue, making the problem tractable.

This equivalence of Eq. \ref{eq:os-MODI} to a Poisson problem with Neumann conditions introduces a conflicting effect. On one hand, it induces a smoothing effect. On the other hand, as \citet{Pan2016, Faiella2021} and \citet{Nie2022} highlighted, the pressure reconstruction is more sensitive to the error located near a Neumann boundary and/or far away from a Dirichlet boundary. It is still not clear how the regularization reconciles with data errors, but this interplay might be related to the disparate outcomes observed with ODI methods in previous studies. Nonetheless, the implementation of Dirichlet conditions, achieved by strategically placing a few pressure transducers in appropriate locations, has been shown to enhance pressure reconstruction \citep{Shanmughan2020, Sakib2024}. In this study, we opt not to enforce any Dirichlet conditions. Instead, we deliberately select a squared domain shape to minimize error propagation \citep{Pan2016}.

\subsection*{Green's Function Integral (GFI)}
\label{sec:Methodology:GFI}

The GFI approach proposed by \citet{Wang_2023} is a classical result from potential theory. It begins by considering the conservative vector field $p \nabla G$, which describes how the pressure field responds to changes in the influence of a point source in $\textbf{x}'$, as described by the Green's function $G(\textbf{x}, \textbf{x}')$ that satisfies the Poisson equation. By employing Green's first identity and incorporating the functional representation of pressure, an expression to compute the pressure field is derived (refer to \citet{Wang_2023} for comprehensive details). Following numerical discretization, the expression manifests as:
%
\begin{align}
    p_k &= \frac{1}{(D-1) \pi} \sum_{j=1}^{N_V} \left[ \left( \nabla p \right)_j \cdot \frac{\textbf{r}_{kj}}{r^{D}_{kj}} \right] \delta V \notag \\
    &\quad - \frac{1}{(D-1) \pi} \sum_{k'=1}^{N_b} p_{k'} \left( \frac{\textbf{r}_{kk'}}{r^{D}_{kk'}} \cdot \delta S_{k'} \right) \mbox{ ,} \label{eq:BEM_boundary}
\end{align}
where the indices $k$ and $k'$ stand for the boundary elements and $j$ corresponds to the center of the cells in the fluid volume. The term $\textbf{r} / r^D$ is the scaled distance between the indicial points, in which $D$ is the dimension of the problem (i.e., $D=2$ in two dimensions and $D=3$ in three dimensions). Finally, the summation limits $N_V$ and $N_b$ refer to the number of volume cells and boundary elements, respectively.

After solving the linear system \ref{eq:BEM_boundary}, the pressure $p_i$ at the $i$ inner nodal points is computed as:
%
\begin{align}
    p_i &= \frac{1}{2 (D-1) \pi} \sum_{j=1}^{N_V} \left[ \left(\nabla p \right)_j \cdot \frac{\textbf{r}_{ij}}{r^{D}_{ij}} \right] \delta V \notag \\
    &\quad - \frac{1}{2 (D-1) \pi} \sum_{k=1}^{N_b} p_{k} \left( \frac{\textbf{r}_{ik}}{r^{D}_{ik}} \cdot \delta S_{k} \right) \mbox{ .}
\end{align}
%

\end{document}